# Decentralized Gossip Learning and Federated Averaging for Histopathology Image Classification

Yusuf Ozturk[1], Bengisu Atli[1,2], Enes Goktekin[2], Akin Ozturk[3], Ulas Bagci[4]

[1] Antalya Bilim University, Electrical and Electronics Engineering Dep. Antalya, Türkiye
[2] Antalya Bilim University, Computer Engineering Dep. Antalya, Türkiye
[3] Ankara University, Graduate School of Natural and Applied Sciences, Ankara, Türkiye
[4] Northwestern University, Dept. of Radiology, Chicago, IL, USA

## Abstract

Breast histopathology analysis increasingly relies on distributed learning because direct data pooling across institutions is often restricted by privacy, governance, and communication constraints. This study compares server-based Federated Averaging (FedAvg), fully decentralized gossip learning, and Hybrid Gossip–FedAvg for invasive ductal carcinoma (IDC) patch classification. Experiments used 277,524 color image patches with patient-disjoint training, validation, and test partitions and a workload-balanced, Dirichlet-guided allocation across six nodes. Ring, random degree-3, and fully connected gossip topologies were evaluated together with sensitivity analyses for statistical heterogeneity, mixing coefficient, learning rate, model drift, prediction disagreement, calibration, clinically motivated operating points, communication payload, and patient-level IDC burden, together with auxiliary backbone robustness analyses. In the principal $\alpha = 0.3$ experiment, Hybrid Gossip–FedAvg achieved a test area under the receiver operating characteristic curve (ROC-AUC) of 0.8811, closely followed by FedAvg at 0.8801 and fully connected gossip at 0.8751. Across three independent patient-level repetitions, FedAvg and Hybrid Gossip–FedAvg obtained the same mean ROC-AUC of 0.9082, with standard deviations of 0.0037 and 0.0043, respectively. Hybrid achieved the highest mean area under the precision–recall curve of 0.8240, whereas FedAvg produced the lowest mean Brier score of 0.1335. Denser gossip graphs improved discrimination but increased theoretical model payload, while ring gossip remained sensitive to learning rate and mixing strength. Overall, FedAvg provided the most consistently reliable server-based baseline, topology-aware gossip offered a viable decentralized alternative, and Hybrid Gossip–FedAvg provided a balanced compromise between peer-to-peer diffusion and periodic global coordination.

**Keywords:** federated learning; gossip learning; decentralized learning; histopathology image classification; invasive ductal carcinoma; distributed deep learning.

## 1. Introduction

Breast cancer most commonly presents as invasive ductal carcinoma (IDC) or invasive lobular carcinoma (ILC), with IDC representing the predominant invasive subtype [1]. IDC and ILC differ in their clinicopathological characteristics, imaging appearance, and clinical outcomes [2]. Histopathological assessment remains essential for diagnosis, grading, prognosis, and treatment planning; however, variability in specimen interpretation can affect diagnostic agreement among pathologists [3]. In addition, IDC exhibits substantial morphological, molecular, and clinicopathological heterogeneity [4–6], which can complicate consistent manual assessment. Deep-learning techniques, particularly convolutional neural networks (CNNs), have been widely investigated for breast cancer detection and histopathology image classification. CNN-based models can learn discriminative representations directly from image patches and reduce dependence on handcrafted feature engineering [7]. Previous studies have demonstrated the potential of deep learning and shallow–deep CNN architectures for automated analysis of breast histopathology images [8,9].

Despite this progress, most deep-learning models are developed in centralized settings, where images from different sources are pooled within a single computational environment. Large-scale distributed training introduces important computational and communication challenges [10,11]. In medical applications, direct cross-institutional pooling may also be restricted by data-governance requirements, institutional agreements, privacy considerations, storage constraints, and the cost of transferring large image collections. Federated and decentralized learning have therefore emerged as promising approaches for collaborative medical artificial intelligence while maintaining raw-data locality [12,13]. Federated learning enables participating institutions to train a shared model without directly exchanging their raw data. General federated-learning frameworks have identified statistical heterogeneity, communication efficiency, privacy, and system coordination as central challenges [14,15]. Communication-efficient federated strategies reduce the frequency or size of model exchanges between clients and the coordinating server [16]. Among these methods, Federated Averaging (FedAvg) remains the most widely used protocol: a central server distributes a global model, participating clients perform local optimization, and the returned models are combined through weighted averaging [17]. Recent studies have investigated differential privacy for breast cancer diagnosis, convergence under limited local data, and model-heterogeneous federated learning [18–20]. Federated learning has also been evaluated for multi-institutional medical imaging and electronic health-record modelling [21,22]. These studies demonstrate the potential of data-local collaborative training; nevertheless, FedAvg remains dependent on a central coordinating server and synchronous client–server communication. Moreover, retaining raw images locally does not by itself provide a formal privacy guarantee because information may still be inferred from transmitted model parameters or updates.

Gossip learning provides a fully decentralized alternative in which nodes exchange model information only with topology-defined neighbours. Model information propagates through repeated local-training and peer-to-peer mixing steps rather than through global server aggregation. Previous studies have compared gossip learning with conventional federated learning, examined decentralized personalization in medical imaging, and investigated gossip learning on fully distributed data [23–25]. Privacy-preserving and secure learning mechanisms remain complementary to such decentralized architectures because parameter exchange alone does not eliminate the possibility of information leakage [26]. The behaviour of decentralized learning depends not only on local data distributions but also on graph connectivity, communication frequency, and local optimization dynamics. Decentralized stochastic-gradient methods can perform competitively under suitable network and optimization conditions [27], while local stochastic-gradient descent can reduce communication requirements by performing multiple optimization steps between model exchanges [28]. However, sparse peer-to-peer communication can also slow model diffusion and increase variation among node-specific models under strongly non-IID data.

Motivated by these considerations, this study compares fully decentralized gossip learning, server-based FedAvg, and a Hybrid Gossip–FedAvg protocol for IDC patch classification using the Breast Histopathology Images dataset. Patient-disjoint training, validation, and test partitions are used to prevent patient-level data leakage, and Dirichlet-based allocation is employed to simulate different levels of statistical heterogeneity across six training nodes. Gossip learning is evaluated using ring, random degree-3, and fully connected communication topologies.

The experimental framework further examines the effects of the gossip mixing coefficient, learning rate, validation-based checkpoint selection, model drift, and node-level prediction disagreement. The Hybrid Gossip–FedAvg protocol combines peer-to-peer parameter diffusion with periodic weighted server aggregation. Performance is evaluated using ROC-AUC, PR-AUC, accuracy, F1-score, sensitivity, specificity, balanced accuracy, Brier score, expected calibration error, and communication volume. In addition, clinically motivated operating thresholds and patch-derived patient-level IDC burden are analysed. Through this controlled comparison, the study aims to characterize the predictive, calibration, stability, and communication trade-offs among centralized aggregation, decentralized parameter diffusion, and periodic hybrid coordination.

## 2. Background and related work

Distributed deep-learning systems require participating processes to exchange gradients, parameters, or complete model states. As model size and the number of participants increase, communication can become a substantial component of the total computational cost. Gradient compression and quantization have therefore been proposed to reduce the number of transmitted bits. Quantized stochastic-gradient descent, for example, compresses gradient information while retaining convergence guarantees under specified assumptions [29]. Communication compression is not implemented in the present experiments, but it represents a complementary strategy for reducing the bandwidth requirements of federated and gossip-based learning. Maintaining raw data at the participating institutions reduces the need for direct data transfer but does not provide complete protection against leakage from model updates. Privacy-preserving aggregation methods have therefore been developed to limit the exposure of client information during federated optimization [30]. Secure aggregation allows a server to recover an aggregate update without directly observing each participating client's individual contribution [31]. These mechanisms are complementary to the communication protocols investigated in this work. Statistical heterogeneity is another major challenge in federated learning. FedProx modifies local optimization through a proximal term intended to reduce instability when client distributions and computational resources differ [32]. Federated medical-imaging studies have demonstrated collaborative model training for tasks such as brain-tumour segmentation without directly pooling institutional image collections [33]. Decentralized Federated Averaging extends this concept by replacing repeated centralized coordination with peer-to-peer communication [34]. The effects of non-IID client data on federated optimization have also been investigated extensively [35], while practical studies have discussed implementation and deployment considerations for collaborative model training [36]. In decentralized learning, the communication graph determines how rapidly local information propagates through the network. Performing multiple gossip steps can improve agreement among nodes, although each additional exchange increases communication cost [37]. GossipGrad replaces conventional central aggregation with asynchronous or peer-to-peer gradient communication and was developed to improve the scalability of distributed deep learning [38]. Resource heterogeneity and client availability can further affect participation and convergence, motivating resource-aware client-selection strategies [39]. Differentially private federated learning introduces an additional privacy layer but may produce a trade-off between privacy protection and predictive performance [40].

In healthcare, the use and disclosure of protected health information are governed by regulatory and institutional requirements. In the United States, the Health Insurance Portability and Accountability Act establishes conditions for handling protected health information but does not categorically prohibit collaborative or distributed research [41]. Federated learning has consequently been investigated across healthcare informatics and medical applications as a means of enabling collaborative model development while maintaining data locality [42,43]. Decentralized federated learning has also been evaluated for healthcare-network applications, including tumour segmentation [44]. Application-oriented studies have explored federated CNN models for multi-institutional disease classification [45], privacy-oriented hybrid federated frameworks for intelligent healthcare systems [46], and federated learning for breast histopathology classification [47]. These studies demonstrate the potential of collaborative learning in medical imaging, but most employ server-based aggregation and focus primarily on predictive performance. Fully decentralized gossip learning remains less extensively studied in computational pathology. In particular, the combined effects of non-IID severity, graph topology, model mixing, node-level disagreement, probability calibration, and communication volume have received limited joint evaluation. Direct comparisons of FedAvg, pure gossip learning, and periodic hybrid aggregation under identical patient partitions and local optimization settings are also scarce.

The present study addresses this gap through a controlled comparison of FedAvg, topology-aware gossip learning, and Hybrid Gossip–FedAvg for IDC patch classification. Unlike evaluations restricted to a single non-IID configuration or a single communication graph, the experiments consider multiple Dirichlet heterogeneity levels and ring, random degree-3, and fully connected gossip topologies. The analysis further includes mixing-coefficient and learning-rate sensitivity, validation-based checkpoint selection, parameter drift, node-level prediction disagreement, probability calibration, clinically motivated operating points, communication volume, and patch-derived patient-level IDC burden.

Accordingly, the contribution of this study is not limited to comparing classification accuracy. The proposed evaluation framework jointly considers discrimination, calibration, model consistency, patient-level aggregation, and system-level communication behaviour. The objective is to identify the conditions under which centralized, decentralized, and hybrid aggregation provide different trade-offs rather than to claim that one protocol is universally superior.

# 3. Methods

## 3.1. Gossip learning

Gossip learning is a decentralized training protocol in which each node performs local optimization using its own data and exchanges model states only with topology-defined neighboring nodes. Unlike server-based Federated Averaging, no central aggregation server participates in the gossip update. The protocol therefore removes reliance on global server aggregation, although fault tolerance under node or communication failures is not explicitly evaluated in this study. This section describes the local optimization objective, parameter-mixing rule, communication topologies, synchronization procedure, and communication-cost calculation used in the experiments.

Let $D_i$ denote the local dataset assigned to node $i$, and let $\mathbf{w}_i^{(r)}$ denote the model state of node $i$ at the beginning of communication round $r$. Each node minimizes a local empirical objective based on binary cross-entropy:

$$F_i(\mathbf{w}) = \mathbb{E}_{(x,y)\sim D_i}[\ell_i(\mathbf{w}; x, y)], \tag{1}$$

where $\ell_i(\cdot)$ denotes the binary cross-entropy loss, including the node-specific class weighting described in Section 3.1.2.

At communication round $r$, node $i$ first performs local optimization starting from $\mathbf{w}_i^{(r)}$. Let $\mathbf{u}_i^{(r)} = \mathrm{LocalTrain}\left(\mathbf{w}_i^{(r)}, D_i\right)$ denote the locally updated model after the prescribed number of local epochs. Consensus is not imposed through an additional regularization term. Instead, each node explicitly mixes its locally updated model with the models received from the nodes in its neighbor set $\mathcal{N}(i)$. The general gossip update is

$$\mathbf{w}_i^{(r+1)} = (1-\beta)\mathbf{u}_i^{(r)} + \frac{\beta}{|\mathcal{N}(i)|}\sum_{j\in\mathcal{N}(i)} \mathbf{u}_j^{(r)}, \tag{2}$$

where $\beta \in [0,1]$ is the gossip mixing coefficient. A value of $\beta = 0$ preserves the locally trained model without neighbor mixing, whereas $\beta = 1$ replaces the local model with the mean of the neighboring models. To assess the sensitivity of gossip learning to the strength of parameter mixing, we evaluate $\beta \in \{0.25, 0.50, 0.6667, 1.00\}$. For the six-node ring topology, each node communicates with its two immediate neighbors. Equation (2) therefore becomes

$$\mathbf{w}_i^{(r+1)} = (1-\beta)\mathbf{u}_i^{(r)} + \frac{\beta}{2}\left(\mathbf{u}_{i-1}^{(r)} + \mathbf{u}_{i+1}^{(r)}\right), \tag{3}$$

where the node indices are taken modulo $N$. For $\beta = 2/3$, the coefficients of the locally updated model and the two neighboring models are each equal to $1/3$, reproducing equal averaging among the three model states. In addition to the ring topology, random degree-3 and fully connected gossip graphs are evaluated. The ring graph contains six undirected edges and provides sparse neighbor exchange, whereas the random degree-3 graph contains nine undirected edges and provides an intermediate level of connectivity. The fully connected graph contains $N(N-1)/2=15$ undirected edges for $N=6$, allowing every node to communicate with all other nodes. The three communication structures used in the experiments are illustrated in Fig. 1. The random graph is generated before training and remains fixed throughout the corresponding experiment. These topologies enable a controlled evaluation of the relationship between graph connectivity, predictive performance, node agreement, and communication volume. The gossip protocol is implemented using synchronous communication rounds. In each round, all nodes first perform the same number of local epochs. The nodes then exchange their locally updated model states with their connected neighbors and apply Eq. (2). The exchanged state includes the trainable model parameters and the Batch Normalization running statistics. No server-based aggregation is performed during gossip training.

Let $\mathcal{E}$ denote the undirected edge set of the communication graph and let $M$ denote the number of float32 values contained in one transmitted model state. Since each float32 value occupies four bytes and each undirected edge produces two directed transmissions per round, the total model payload transmitted over $R$ rounds is

$$C_{\text{Gossip}}(\text{MB}) = \frac{2|\mathcal{E}|\,(4M)\,R}{10^6}. \quad (4)$$

The factor of 2 accounts for bidirectional transmission over each edge. Accordingly, the ring, random degree-3, and fully connected graphs require 12, 18, and 30 model transmissions per round, respectively. Communication cost is reported in decimal megabytes, where $1\,\text{MB} = 10^6$ bytes. The calculation represents the model payload and excludes protocol headers, MPI control messages, and other implementation-dependent overhead. The reported experiments use $N = 6$ training nodes and a maximum of $R = 5$ communication rounds. Model checkpoints are compared using validation ROC-AUC, and the checkpoint with the highest validation ROC-AUC is subsequently evaluated on the held-out test set. The test set is not used for checkpoint selection or hyperparameter tuning.

To examine stability across communication rounds, node-specific models are compared with the arithmetic mean model $\bar{\mathbf{w}}^{(r)} = \frac{1}{N}\sum_{i=1}^{N} \mathbf{w}_i^{(r)}$. Parameter drift is quantified using the Euclidean distance and relative Euclidean distance between each node model and $\bar{\mathbf{w}}^{(r)}$, together with their cosine similarity. Prediction disagreement is assessed by applying the node-specific models to the same evaluation samples and measuring the dispersion of predicted probabilities and the proportion of samples for which the node-level binary decisions are not unanimous. These analyses are complemented by the learning-rate and mixing-coefficient ablations.

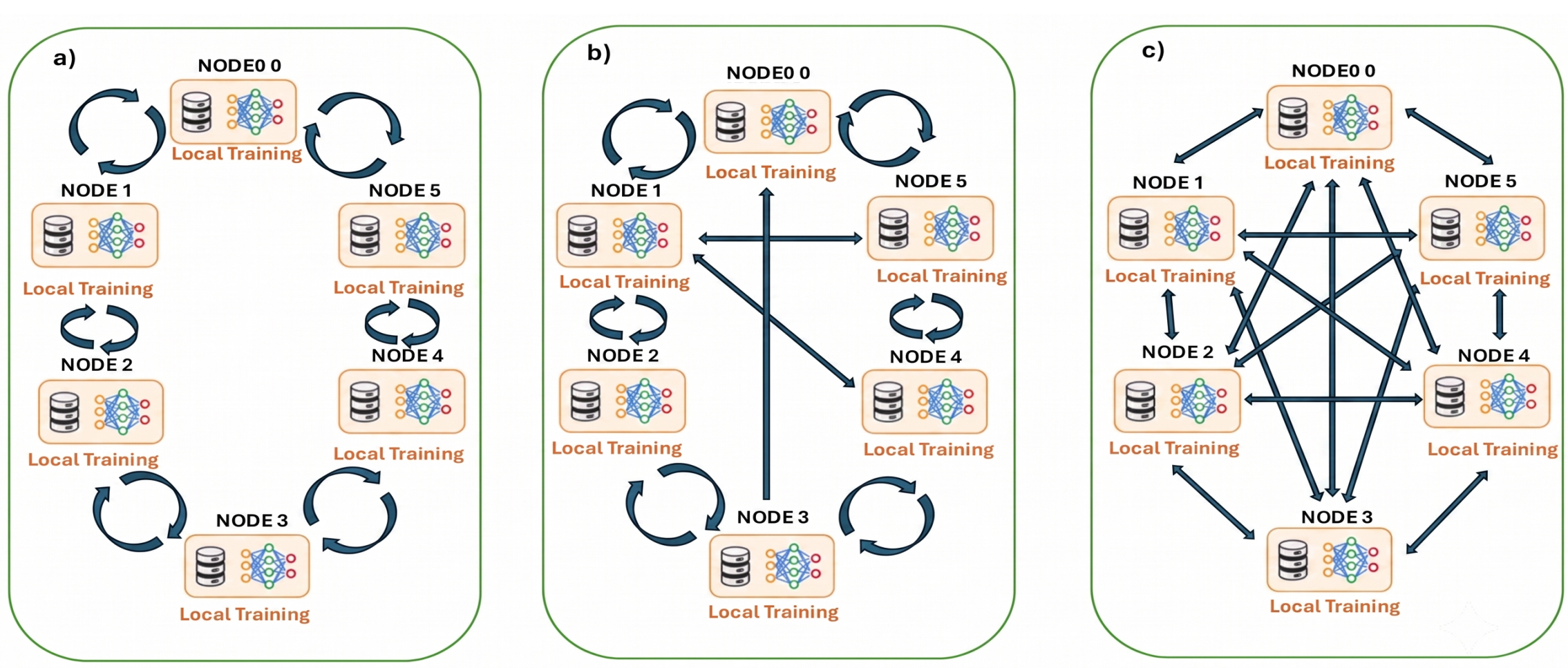

**Fig. 1** Six-node gossip communication topologies used in the experiments: (a) ring, (b) random degree-3, and (c) fully connected. Each node performs local optimization using its private data and subsequently exchanges model states with its topology-defined neighbors

As shown in Fig. 1, increasing graph connectivity provides more direct paths for model-state propagation across nodes communication rounds but also increases the number of model transmissions per communication round. These three topologies are therefore used to examine the trade-off between predictive performance, node-level agreement, and communication cost.

#### *3.1.1.* Dataset and non-IID partitioning

We used the publicly available Breast Histopathology Images dataset for invasive ductal carcinoma classification [48]. The source dataset was derived from 162 whole-slide breast histopathology images acquired at 40 × magnification and released as 277,524 RGB image patches of size 50 × 50 pixels. The dataset contains 198,738 IDC-negative and 78,786 IDC-positive patches, corresponding to an overall IDC-positive patch prevalence of approximately 28.4%. The extracted archive used in our experiments contained 279 unique top-level case identifiers. Each directory identifier was treated as the patient-level grouping variable, and all patches associated with the same identifier were kept together during partitioning. This grouping strategy prevents patches originating from the same case from appearing in more than one experimental subset. The case identifiers were randomly divided into training, validation, and held-out test partitions using proportions of 70%, 10%, and 20%, respectively. The training partition was used for local model optimization, whereas the validation partition was used for checkpoint selection and, where applicable, operating-threshold determination. The held-out test partition was reserved for final evaluation. No patches from the test partition were used for model selection, hyperparameter tuning, calibration fitting, or threshold selection.

The training cases were subsequently distributed across $N = 6$ nodes using a workload-balanced, Dirichlet-guided allocation procedure. For each case $p$, the dominant patch label was defined as $c_p = \arg\max_{c\in\{0,1\}} n_{p,c}$, where $n_{p,c}$ denotes the number of patches of class $c$belonging to case $p$. For each dominant class $c$, a node-preference vector was sampled as $\boldsymbol{\pi}_c \sim \mathrm{Dirichlet}(\alpha\mathbf{1}_N)$, where $\alpha$ controls the concentration of the class-dependent node preferences. Smaller values of $\alpha$generally produce more concentrated preferences and therefore tend to increase differences in class prevalence among nodes. Because the number of patches associated with individual cases varies substantially, an unconstrained Dirichlet allocation may generate empty nodes or severe workload imbalance. Cases were therefore processed in descending order of patch count and assigned using a combined allocation score. For each candidate node, 70% of the score represented its remaining workload capacity and 30% represented the Dirichlet-derived preference for the dominant label of the current case. The target workload was defined as the mean number of training patches per node, and a soft upper limit of 1.35 times this target was used to discourage excessive concentration at a single node.

This allocation procedure preserves case-level exclusivity and introduces controlled client-level statistical heterogeneity while maintaining computationally feasible workloads. Accordingly, $\alpha$ controls the Dirichlet component of the allocation rather than defining an unconstrained patch-level Dirichlet partition. The principal experiments used $\alpha = 0.3$, whereas sensitivity to statistical heterogeneity was examined using $\alpha \in \{0.1, 0.3, 0.5, 1.0\}$. For every matched comparison, FedAvg, gossip learning, and Hybrid Gossip–FedAvg used the same training, validation, and test partitions and the same node-level case assignments. This experimental design isolates the effects of communication topology and aggregation strategy from variation caused by different data partitions. The principal topology, heterogeneity, mixing-coefficient, and learning-rate experiments were conducted using a fixed partition and initialization setting. Robustness was additionally evaluated for the three principal protocols—ring-topology gossip learning, FedAvg, and Hybrid Gossip–FedAvg—using three independent repetitions with seeds 42, 123, and 2026. For each repetition, the complete case-level splitting and node-allocation process was regenerated, while all three protocols within the repetition used identical partitions and node assignments.

#### *3.1.2.* Local dataset preparation

For each node $i$, the local dataset $D_i$ consisted exclusively of patches associated with the training cases assigned to that node. The same preprocessing procedure was applied across all learning protocols. Image intensities were normalized before model input, and lightweight augmentation was applied only during local training using random horizontal and vertical flips. No augmentation was applied to the validation or held-out test partitions. This strategy increased local sample diversity without changing class labels or transferring patches across patient-level partitions.

Because the IDC-positive prevalence varied among node-specific training shards, class weights were calculated independently for each node. Let $n_i$ denote the total number of training patches at node $i$, and let $n_{i,c}$ denote the number belonging to class $c \in \{0,1\}$. The class weight was defined as $\omega_{i,c} =$ $n_i/(2n_{i,c})$, where the denominator factor of two corresponds to the number of classes. These weights approximately balanced the contribution of IDC-positive and IDC-negative patches to the local binary cross-entropy loss without modifying the underlying patient or patch distribution. The same preprocessing, augmentation, and node-specific weighting procedures were used for gossip learning, FedAvg, and Hybrid Gossip–FedAvg so that differences in performance could be attributed primarily to their communication and aggregation mechanisms.

### *3.1.3.* CNN architecture

All six nodes used the same lightweight CNN so that differences among gossip learning, FedAvg, and Hybrid Gossip–FedAvg could be attributed to the communication and aggregation protocols rather than to model capacity. The network received an RGB histopathology patch $\mathbf{x} \in \mathbb{R}^{50\times50\times3}$ and contained three sequential convolutional blocks. The first block used 32 $3 \times 3$ filters with same padding, He initialization, $L_2$ kernel regularization, Batch Normalization, LeakyReLU activation with a negative-slope coefficient of 0.01, $2 \times 2$ max pooling, and dropout of 0.15. The second block followed the same structure with 64 filters and dropout of 0.25, whereas the third block used 128 filters followed by Batch Normalization and LeakyReLU without an additional pooling layer. The resulting feature maps were passed through global average pooling and dropout of 0.40, followed by a single sigmoid-activated output unit that produced the estimated probability $p(y = 1 \mid \mathbf{x})$ of an IDC-positive patch.

The architectural and baseline optimization settings are summarized in Fig. 2. Unless otherwise stated, the convolutional kernels used an $L_2$ regularization coefficient of $1 \times 10^{-4}$, and the model was optimized using binary cross-entropy. The same initialization procedure, architecture, and output definition were retained across all principal protocol comparisons. Two auxiliary ResNet-18 analyses were conducted to examine whether the reported findings were specific to the lightweight CNN. The first evaluated ring-topology gossip under the principal five-round setting, whereas the second evaluated FedAvg using a longer validation-controlled communication budget. These auxiliary experiments were not used to replace the shared lightweight CNN in the primary cross-protocol comparison.

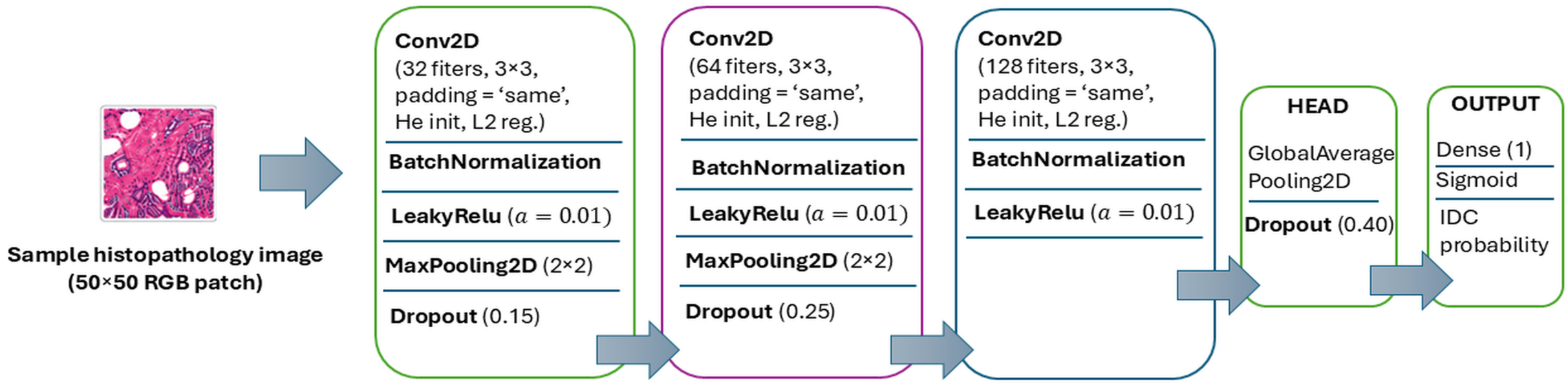

**Fig. 2** CNN architecture used for IDC patch classification and shared across all distributed learning protocols

### *3.1.4.* Experimental setup and evaluation protocol

Unless otherwise stated, all principal experiments used $N = 6$ training nodes, $R = 5$ scheduled communication rounds, one local epoch per round, and a batch size of 32. Local optimization was performed using Adam with a baseline learning rate of $1 \times 10^{-3}$; alternative learning rates were examined only in the dedicated sensitivity analysis. The principal non-IID setting used $\alpha = 0.3$, while $\alpha \in \{0.1, 0.3, 0.5, 1.0\}$ was evaluated to examine sensitivity to client heterogeneity. Gossip learning was tested using ring, random degree-3, and fully connected communication graphs. The Hybrid Gossip–FedAvg protocol used ring-based gossip mixing with periodic server aggregation every $H = 2$ rounds. All matched protocol comparisons used identical case-level partitions, node assignments, CNN architecture, initialization policy, local epoch count, batch size, and preprocessing procedure.

Because gossip learning maintains a separate model at each node, a post-hoc arithmetic mean model was constructed after every communication round to enable comparison with the single global model produced by FedAvg. For round $r$, this reporting model was defined as

$$\bar{\mathbf{w}}^{(r)} = \frac{1}{N}\sum_{i=1}^{N} \mathbf{w}_i^{(r)} . \quad (5)$$

The averaging operation was used only for centralized evaluation and did not participate in gossip training or alter subsequent node updates. Unless otherwise specified, the reported global gossip metrics correspond to this post-hoc averaged model. Node-specific models were additionally retained for the drift and prediction-disagreement analyses.

For the principal experiments, all five scheduled communication rounds were completed, and the checkpoint with the highest validation ROC-AUC was selected for final evaluation on the held-out test partition. The test partition was not used to select rounds, tune hyperparameters, fit calibration parameters, or determine clinical operating thresholds. Accordingly, this procedure is described as validation-based checkpoint selection, rather than early stopping. In the three independent robustness repetitions, training was allowed to terminate before the fifth round when validation ROC-AUC failed to improve for two consecutive rounds; the best validation checkpoint was retained in every case. This distinction ensures that the single-split sensitivity experiments and the independently repeated robustness experiments are reported using terminology consistent with their actual stopping procedures.

An additional extended-budget FedAvg experiment used a ResNet-18 backbone trained from scratch. This auxiliary experiment used a fixed patient-disjoint partition comprising 197 training, 27 validation, and 55 held-out test case identifiers, together with the corresponding six client assignments. Each communication round comprised one complete local epoch at every client, using a batch size of 128 and Adam with a learning rate of $1 \times 10^{-4}$. Training was permitted to continue for a maximum of 20 communication rounds and employed validation ROC-AUC early stopping with a patience of five evaluated rounds, a minimum training duration of ten rounds, and a minimum improvement criterion of 0.001. Training terminated after 16 completed rounds, and the checkpoint with the highest validation ROC-AUC was obtained at round 13. The held-out test partition was evaluated only once after restoration of this validation-selected checkpoint. The experiment was implemented as a single-process, single-GPU simulation and was treated as an auxiliary backbone-and-budget analysis rather than as part of the matched primary protocol comparison.

The principal distributed experiments were implemented in Python using mpi4py with OpenMPI communication over Ethernet and were executed on a workstation equipped with a 12th-generation Intel Core i5-12500H processor, 16 GB of RAM, and an NVIDIA GeForce RTX 3050 Ti Laptop GPU. The extended ResNet-18 FedAvg experiment was conducted separately as a single-process, single-GPU simulation using Kaggle computational resources. Accordingly, the ResNet-18 experiment was used to evaluate backbone capacity and extended communication-round behavior rather than implementation-level runtime or network latency. Communication costs for all experiments were therefore compared using theoretical float32 model payloads rather than measured wall-clock communication times.

#### *3.1.5.* Evaluation metrics

Model performance was evaluated at both fixed and variable decision thresholds. At the fixed threshold $t = 0.5$, we computed accuracy, precision, sensitivity, specificity, balanced accuracy, and F1-score. Accuracy was calculated as $(TP + TN)/(TP + TN + FP + FN)$, and the F1-score as $2(\text{precision} \times \text{sensitivity})/(\text{precision} + \text{sensitivity})$, where $TP$, $TN$, $FP$, and $FN$ denote true positives, true negatives, false positives, and false negatives, respectively. Threshold-independent discrimination was assessed using ROC-AUC and PR-AUC. ROC-AUC summarizes the relationship between sensitivity and false-positive rate across decision thresholds, whereas PR-AUC summarizes the precision–recall relationship and is particularly informative for the imbalanced IDC-positive class.

Probabilistic reliability was evaluated using the Brier score and expected calibration error (ECE). For $n$ patches with binary labels $y_j$ and predicted IDC-positive probabilities $p_j$, the Brier score was calculated as $n^{-1} \sum_{j=1}^{n}(p_j - y_j)^2$. ECE was computed using 15 equally spaced probability bins as the weighted average of the absolute difference between the mean predicted probability and observed positive frequency within each non-empty bin. Lower Brier score and ECE values indicate better probabilistic reliability; these measures were interpreted as calibration metrics rather than discrimination metrics.

Clinically motivated operating points were evaluated by selecting decision thresholds on the validation partition and applying them unchanged to the held-out test partition. Thresholds targeting sensitivities of 90% and 95% were selected by maximizing specificity among validation thresholds satisfying the corresponding sensitivity target. Conversely, thresholds targeting specificities of 90% and 95% were selected by maximizing sensitivity among validation thresholds satisfying the corresponding specificity target. The test partition was not used to determine these thresholds. The final comparison emphasizes specificity at 95% sensitivity and sensitivity at 95% specificity, while the 90% operating points were retained as supporting analyses.

Patient-level IDC burden was evaluated by aggregating patch-level predictions within each held-out case. For patient $p$, the true IDC burden was defined as the proportion of its patches labelled IDC-positive, and the predicted burden was defined as the mean IDC-positive probability across its patches. Agreement between true and predicted burden was quantified using mean absolute error, root-mean-square error, Pearson correlation, and Spearman rank correlation. Binary patient-level classification was not reported because every held-out patient contained at least one IDC-positive patch, making simple patient-level presence/absence labels non-informative. Results from the independent robustness analysis were summarized as the mean and standard deviation across the three patient-level repetitions.

### *3.2.* Federated Averaging (FedAvg)

Federated Averaging (FedAvg) is a server-based federated learning protocol in which participating clients optimize a shared model without transferring their local image data to a central repository [17]. Let $D_k$ denote the local training dataset of client $k$, containing $n_k$ patches, and let $n = \sum_{k=1}^{K} n_k$. The global optimization objective can be written as $F(\mathbf{w}) = \sum_{k=1}^{K}(n_k/n)F_k(\mathbf{w})$, where $F_k$ is the empirical loss evaluated on client $k$. At the beginning of communication round $r$, the server broadcasts the current global model $\mathbf{w}^{(r)}$ to all $K = 6$ clients. Each client then performs the prescribed local optimization starting from this shared model and returns the locally updated state $\mathbf{u}_k^{(r)} = \mathrm{LocalTrain}(\mathbf{w}^{(r)}, D_k)$ to the server.

The server constructs the global model for the next communication round using sample-size-weighted averaging:

$$\mathbf{w}^{(r+1)} = \sum_{k=1}^{K} \frac{n_k}{\sum_{j=1}^{K} n_j} \mathbf{u}_k^{(r)}. \tag{6}$$

The aggregated state includes the trainable parameters and the running statistics of the Batch Normalization layers. FedAvg uses exactly the same case-level training, validation, and test partitions, node assignments, preprocessing, augmentation, node-specific class weighting, CNN architecture, batch size, local epoch count, and baseline optimizer settings as the matched gossip-learning and Hybrid Gossip–FedAvg experiments. Model checkpoint selection and held-out test evaluation follow the validation-based protocol described in Section 3.1.4. Consequently, differences between matched experiments are attributable primarily to the communication and aggregation strategy rather than to differences in data preparation or model capacity.

The server–client communication pattern is illustrated in Fig. 3(a). Each communication round involves $K$ server-to-client model transmissions and $K$ client-to-server transmissions. Let $M$ denote the number of float32 values contained in one complete model state. The theoretical model payload transmitted over $R$ scheduled rounds is therefore

$$C_{\mathrm{FedAvg}}(\mathrm{MB}) = \frac{2K\,(4M)R}{10^6}. \quad (7)$$

The factor of two accounts for one download and one upload per client in each round. Transmission of client sample counts contributes negligible additional traffic compared with the complete model state. As with the gossip communication calculation, the reported volume represents theoretical model payload in decimal megabytes and excludes MPI control traffic, protocol headers, serialization overhead, and other implementation-dependent communication. FedAvg produces a shared global model after every round but requires synchronous communication with an available central server.

### 3.3. Hybrid Gossip–FedAvg

Hybrid Gossip–FedAvg combines peer-to-peer gossip mixing with periodic server-based aggregation. Between global synchronization events, clients exchange model states only with their topology-defined neighbors, as in decentralized gossip learning. Every $H$ communication rounds, the gossip-mixed client models are additionally aggregated by a central server using the FedAvg rule. The protocol therefore preserves local model diffusion between neighboring clients while periodically reducing cross-client divergence through global coordination.

At round $r$, client $i$ first performs one local epoch of Adam optimization and obtains $\mathbf{u}_i^{(r)} = \mathrm{LocalTrain}_{\mathrm{Adam}}(\mathbf{w}_i^{(r)}, D_i)$. Ring-topology gossip mixing then produces the provisional model

$$\widetilde{\mathbf{w}}_i^{(r+1)} = (1-\beta)\mathbf{u}_i^{(r)} + \frac{\beta}{|\mathcal{N}(i)|}\sum_{j\in\mathcal{N}(i)} \mathbf{u}_j^{(r)}. \quad (8)$$

If $r+1$ is not a multiple of $H$, the next-round client model is simply $\mathbf{w}_i^{(r+1)} = \widetilde{\mathbf{w}}_i^{(r+1)}$. At every periodic synchronization round, the server computes a sample-size-weighted average of the gossip-mixed models,

$$\mathbf{w}_{\mathrm{global}}^{(r+1)} = \sum_{i=1}^{K} \frac{n_i}{\sum_{j=1}^{K} n_j} \widetilde{\mathbf{w}}_i^{(r+1)}, \quad (9)$$

and broadcasts the resulting state to all clients, so that $\mathbf{w}_i^{(r+1)} = \mathbf{w}_{\mathrm{global}}^{(r+1)}$ for every $i$. The aggregated state includes both trainable parameters and Batch Normalization running statistics. In the principal Hybrid experiments, $K = 6$, ring-topology gossip is used with $\beta = 2/3$, and global synchronization is performed every $H = 2$ rounds. Over the five scheduled rounds, server aggregation therefore occurs after rounds 2 and 4.

The Hybrid communication pattern is illustrated in Fig. 3(b), while Fig. 3(a) shows the corresponding server-based FedAvg structure. The theoretical Hybrid model payload is the sum of the ring-gossip transmissions over all $R$ rounds and the additional client–server exchanges occurring at the $\lfloor R/H \rfloor$ synchronization events:

$$C_{\mathrm{Hybrid}}(\mathrm{MB}) = C_{\mathrm{Gossip}}(\mathrm{MB}) + \left\lfloor \frac{R}{H} \right\rfloor \frac{2K\,(4M)}{10^6}. \quad (10)$$

As in the preceding communication calculations, this quantity represents float32 model payload only and excludes protocol headers, MPI control traffic, serialization overhead, and evaluation-related communication. Hybrid Gossip–FedAvg therefore requires more communication than ring gossip alone, but less server communication than performing FedAvg aggregation after every round.

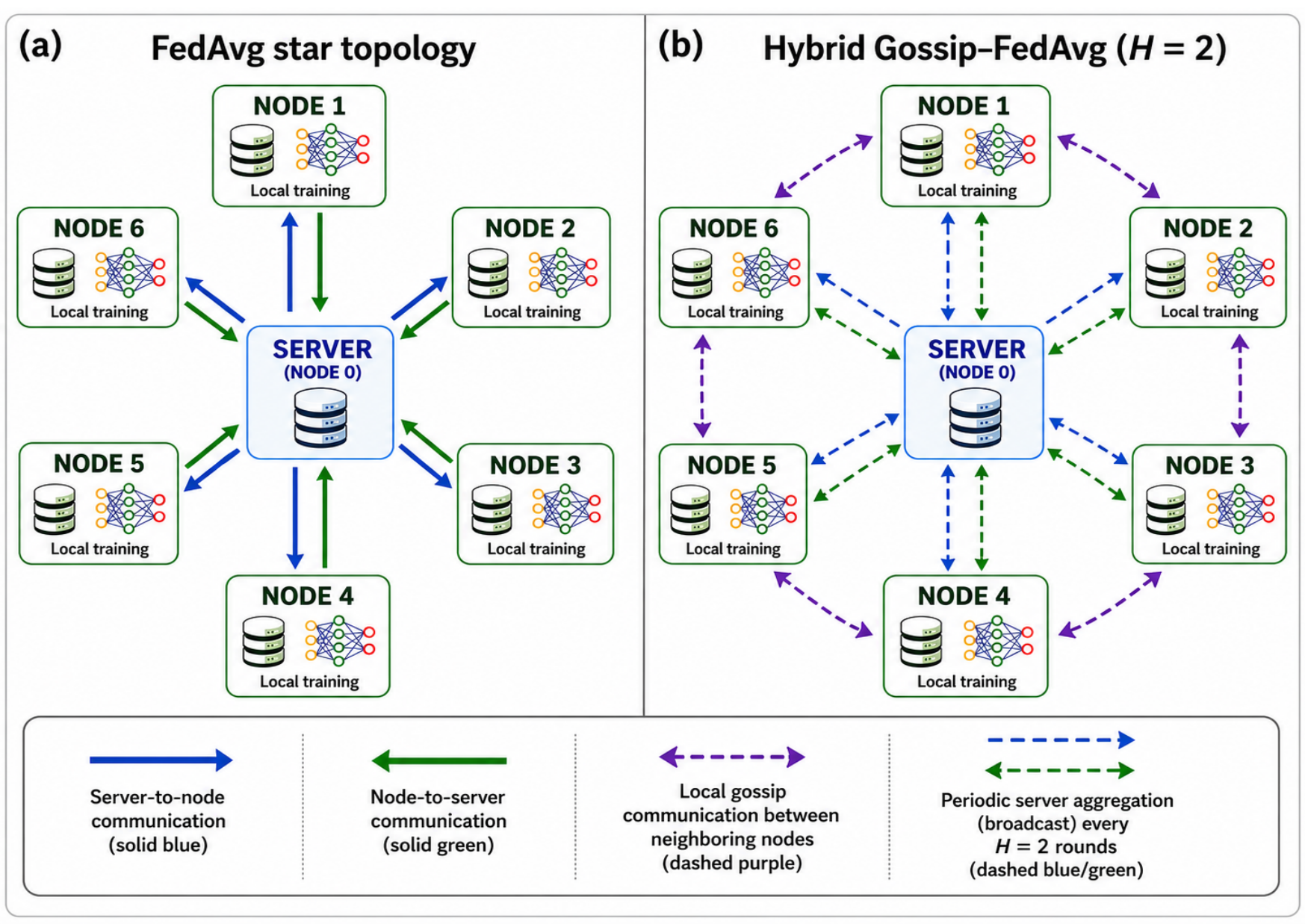

**Fig. 3** Communication structures of (a) server-based FedAvg and (b) Hybrid Gossip–FedAvg with periodic global aggregation every $H = 2$ rounds

### 3.4. Comparative evaluation framework

The three distributed learning protocols were compared across four complementary dimensions. Predictive discrimination was assessed using ROC-AUC, PR-AUC, accuracy, F1-score, sensitivity, specificity, and balanced accuracy. Probabilistic reliability and clinically oriented behavior were evaluated using Brier score, expected calibration error, and validation-selected operating thresholds targeting high sensitivity or high specificity. Training stability was characterized through validation-based checkpoint selection, round-wise performance, parameter drift, cosine similarity, and node-level prediction disagreement. System-level efficiency was evaluated using the theoretical float32 model payload transmitted over the five scheduled communication rounds. Runtime values were treated as descriptive results specific to the reported hardware, software, and network configuration and were not interpreted as hardware-independent protocol latencies.

All matched comparisons used identical case-level partitions, node assignments, preprocessing, CNN architecture, optimizer configuration, batch size, and local epoch count unless an experiment explicitly varied one factor. The principal single-partition analysis was used to compare communication topologies, Dirichlet heterogeneity settings, gossip mixing coefficients, and learning rates. Two auxiliary ResNet-18 analyses were conducted separately: a five-round ring-gossip backbone robustness check and an extended-budget FedAvg experiment examining the combined effects of increased model capacity and a longer validation-controlled aggregation schedule. The independent robustness analysis evaluated ring-topology gossip learning, FedAvg, and Hybrid Gossip–FedAvg across three repeated patient-level partitions and initializations using seeds 42, 123, and 2026. Results from these repetitions were summarized as mean $\pm$ standard deviation. Because only three independent repetitions were available, cross-protocol differences were interpreted descriptively rather than as formal evidence of statistical superiority.

## 4. Results

### 4.1. Primary protocol comparison under $\alpha = 0.3$

The principal comparison evaluated ring, random degree-3, and fully connected gossip learning against FedAvg and Hybrid Gossip–FedAvg under the same workload-balanced, Dirichlet-guided allocation with $\alpha = 0.3$. All methods used identical case-level training, validation, and test partitions, the shared lightweight CNN, one local epoch per communication round, and the baseline Adam learning rate of $1 \times 10^{-3}$. All five scheduled rounds were completed, and the checkpoint obtaining the highest validation ROC-AUC was evaluated once on the held-out test partition.

Table 1 summarizes the resulting discrimination performance. Hybrid Gossip–FedAvg achieved the numerically highest test ROC-AUC of 0.8811 and PR-AUC of 0.6849, followed closely by FedAvg with ROC-AUC of 0.8801 and PR-AUC of 0.6725. The ROC-AUC difference between these two protocols was only 0.0010 and should therefore be interpreted as numerical equivalence in the principal single-partition experiment rather than evidence of Hybrid superiority. Among the fully decentralized methods, fully connected gossip produced the highest ROC-AUC of 0.8751, while random degree-3 gossip achieved ROC-AUC of 0.8678. Ring gossip obtained the lowest primary ROC-AUC of 0.8455.

**Table 1** Primary discrimination performance at the validation-selected checkpoint under the $\alpha = 0.3$ allocation.

| *Learning protocol* | *Communication setting* | *Selected Round* | *Test ROC-AUC* | *Test PR-AUC* |
|---|---|---|---|---|
| *Gossip learning* | Ring topology | 4 | 0.8455 | 0.5820 |
| *Gossip learning* | Random degree-3 topology | 5 | 0.8678 | 0.6651 |
| *Gossip learning* | Fully connected topology | 3 | 0.8751 | 0.6500 |
| *Federated Averaging* | Server-based star topology | 3 | 0.8801 | 0.6725 |
| *Hybrid Gossip–FedAvg* | Ring gossip with $H = 2$ | 4 | **0.8811** | **0.6849** |

Increasing gossip connectivity was associated with improved ROC-AUC relative to the sparse ring topology, although the improvement was not strictly monotonic for PR-AUC: random degree-3 gossip achieved a higher PR-AUC than fully connected gossip. This distinction indicates that topology affected ROC ranking performance and positive-class precision–recall behavior differently. The corresponding communication cost is analysed separately in Section 4.3 so that predictive performance and model-payload efficiency are not conflated in the primary discrimination table.

### 4.2. Round-wise stability and model consistency

Round-wise behavior was examined using the independent robustness experiment with seed 42 as a representative matched repetition. FedAvg, ring-topology gossip learning, and Hybrid Gossip–FedAvg used identical patient partitions and node assignments in this comparison. Figure 4(a) and Fig. 4(b) show validation ROC-AUC and PR-AUC across the five scheduled communication rounds. FedAvg reached its highest validation ROC-AUC of 0.9176 at round 3, ring gossip reached 0.9163 at round 4, and Hybrid Gossip–FedAvg reached 0.9150 at round 3. Validation PR-AUC showed a similar but not identical pattern: the highest values were 0.8449 for FedAvg at round 4, 0.8431 for ring gossip at round 4, and 0.8355 for Hybrid Gossip–FedAvg at round 4. Because checkpoint selection was based on validation ROC-AUC, the selected rounds were 3, 4, and 3 for FedAvg, ring gossip, and Hybrid Gossip–FedAvg, respectively.

The node-consistency analysis for ring gossip is presented in Fig. 4(c) and Fig. 4(d). Mean relative parameter drift decreased from 0.1802 in the first round to 0.1632 in the fifth round, while mean cosine similarity increased from 0.9843 to 0.9869. Prediction disagreement also declined overall: the proportion of evaluation samples with non-unanimous node-level binary decisions decreased from 0.3870 to 0.1610, and the mean standard deviation of node-level predicted probabilities decreased from 0.1067 to 0.0553. These results indicate that the node models became progressively more consistent as information propagated through the ring topology.

However, improved parameter and prediction agreement did not produce monotonic gains in validation discrimination. For example, ring-gossip ROC-AUC peaked at round 4 and decreased slightly at round 5 despite lower parameter drift and prediction dispersion. Thus, late-round performance variation cannot be attributed solely to increasing divergence among node models. The findings instead suggest that distributed optimization under non-IID allocation reflects a combination of local fitting, parameter mixing, and consensus dynamics, supporting the use of validation-based checkpoint selection rather than relying automatically on the final communication round.

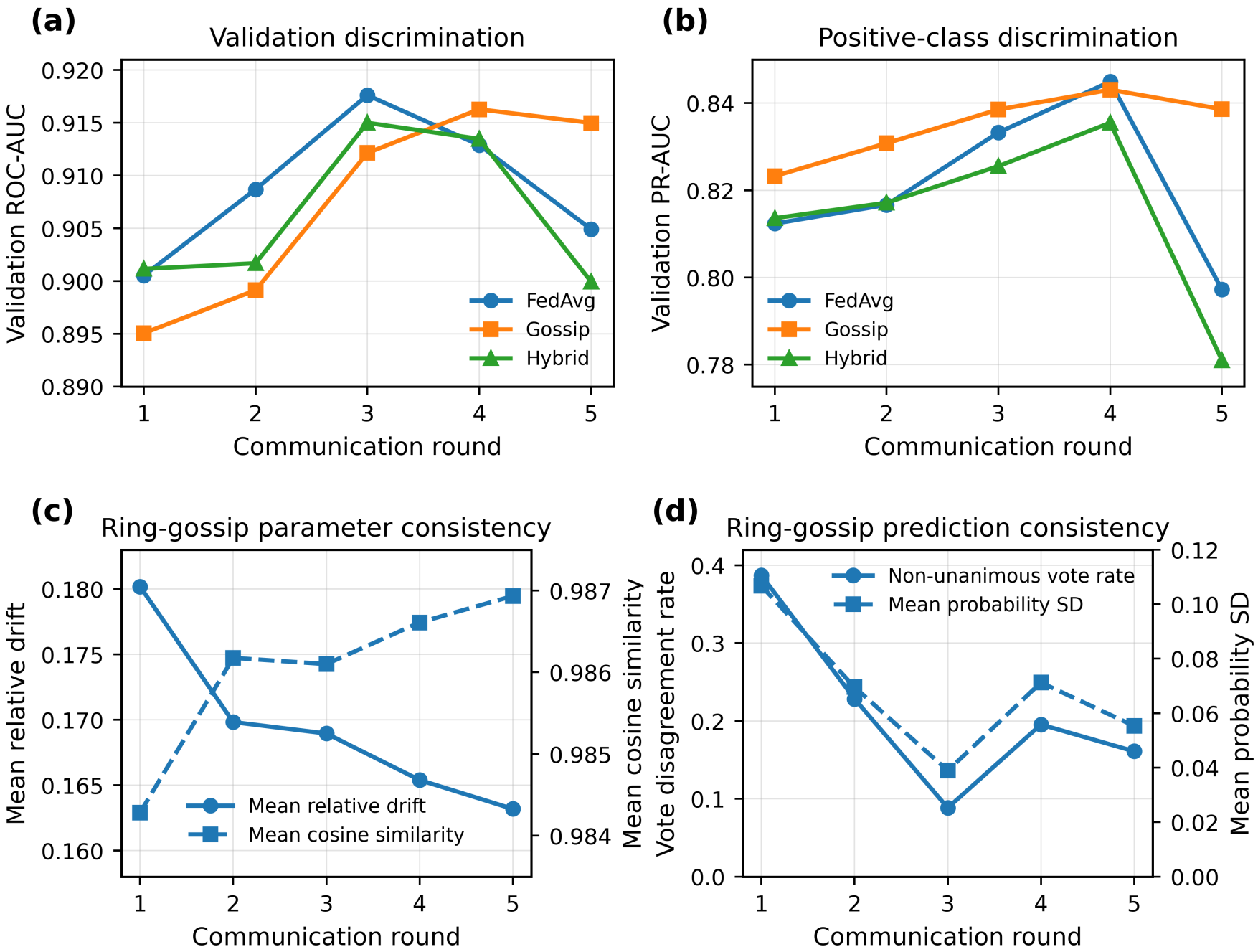

**Fig. 4** Round-wise validation performance and node consistency in the representative seed-42 robustness experiment: (a) validation ROC-AUC, (b) validation PR-AUC, (c) mean relative parameter drift and cosine similarity for ring gossip, and (d) node-level prediction disagreement and mean probability dispersion

### 4.3. Heterogeneity, topology, and communication trade-offs

The sensitivity of FedAvg and ring-topology gossip learning to the heterogeneity-control parameter was evaluated using $\alpha \in \{0.1, 0.3, 0.5, 1.0\}$. As shown in Fig. 5(a), FedAvg remained comparatively stable, with test ROC-AUC values between 0.8801 and 0.8857. Ring gossip showed greater variation, obtaining ROC-AUC values of 0.8619, 0.8455, 0.8675, and 0.8745 for $\alpha = 0.1$, 0.3, 0.5, and 1.0, respectively. The non-monotonic pattern should not be interpreted as a contradiction of Dirichlet heterogeneity theory because $\alpha$controls only the preference component of the workload-balanced allocation. The realized node distributions also depend on case-level patch counts, dominant labels, the load-balancing term, and the specific random allocation.

Communication topology produced a clearer trade-off between predictive performance and transmitted model payload. Table 2 reports the theoretical float32 model traffic accumulated over all five scheduled communication rounds. Ring gossip required 60 directed model transmissions and 22.63 MB of payload, whereas random degree-3 and fully connected gossip required 90 and 150 transmissions, corresponding to 33.94 and 56.56 MB, respectively. Increasing connectivity improved gossip ROC-AUC from 0.8455 for the ring graph to 0.8678 for the random graph and 0.8751 for the fully connected graph. FedAvg required the same theoretical payload as ring gossip, 22.63 MB, because both involved 12 model transmissions per round. Hybrid Gossip–FedAvg combined 60 ring-gossip transmissions with 24 additional client–server transmissions at rounds 2 and 4, resulting in 84 model transmissions and a total theoretical payload of 31.68 MB.

**Table 2** Communication structure, theoretical model payload over five scheduled rounds, and validation-selected test ROC-AUC under the $\alpha = 0.3$ allocation.

| ***Learning protocol*** | ***Communication structure*** | ***Directed model transmissions over five rounds*** | ***Theoretical model payload (MB)*** | ***Validation-selected test ROC-AUC*** |
|---|---|---|---|---|
| *Gossip learning* | Ring topology, 6 edges | 60 | 22.63 | 0.8455 |
| *Gossip learning* | Random degree-3 topology, 9 edges | 90 | 33.94 | 0.8678 |
| *Gossip learning* | Fully connected topology, 15 edges | 150 | 56.56 | 0.8751 |
| *Federated Averaging* | Server-based star topology | 60 | 22.63 | 0.8801 |
| *Hybrid Gossip–FedAvg* | Ring gossip with two FedAvg synchronizations | 84 | 31.68 | 0.8811 |

The communication–performance relationship is visualized in Fig. 5(b). Ring gossip provided the lowest-cost fully decentralized configuration, whereas fully connected gossip achieved the strongest gossip-only ROC-AUC at approximately 2.5 times the ring payload. Random degree-3 gossip occupied an intermediate region, providing a substantial performance gain over ring gossip with less communication than the fully connected graph. FedAvg achieved higher ROC-AUC than all gossip-only configurations at the same theoretical payload as ring gossip, although it required a continuously available central server. Hybrid Gossip–FedAvg achieved the numerically highest ROC-AUC with an intermediate payload, but its 0.0010 ROC-AUC margin over FedAvg is too small to support a superiority claim.

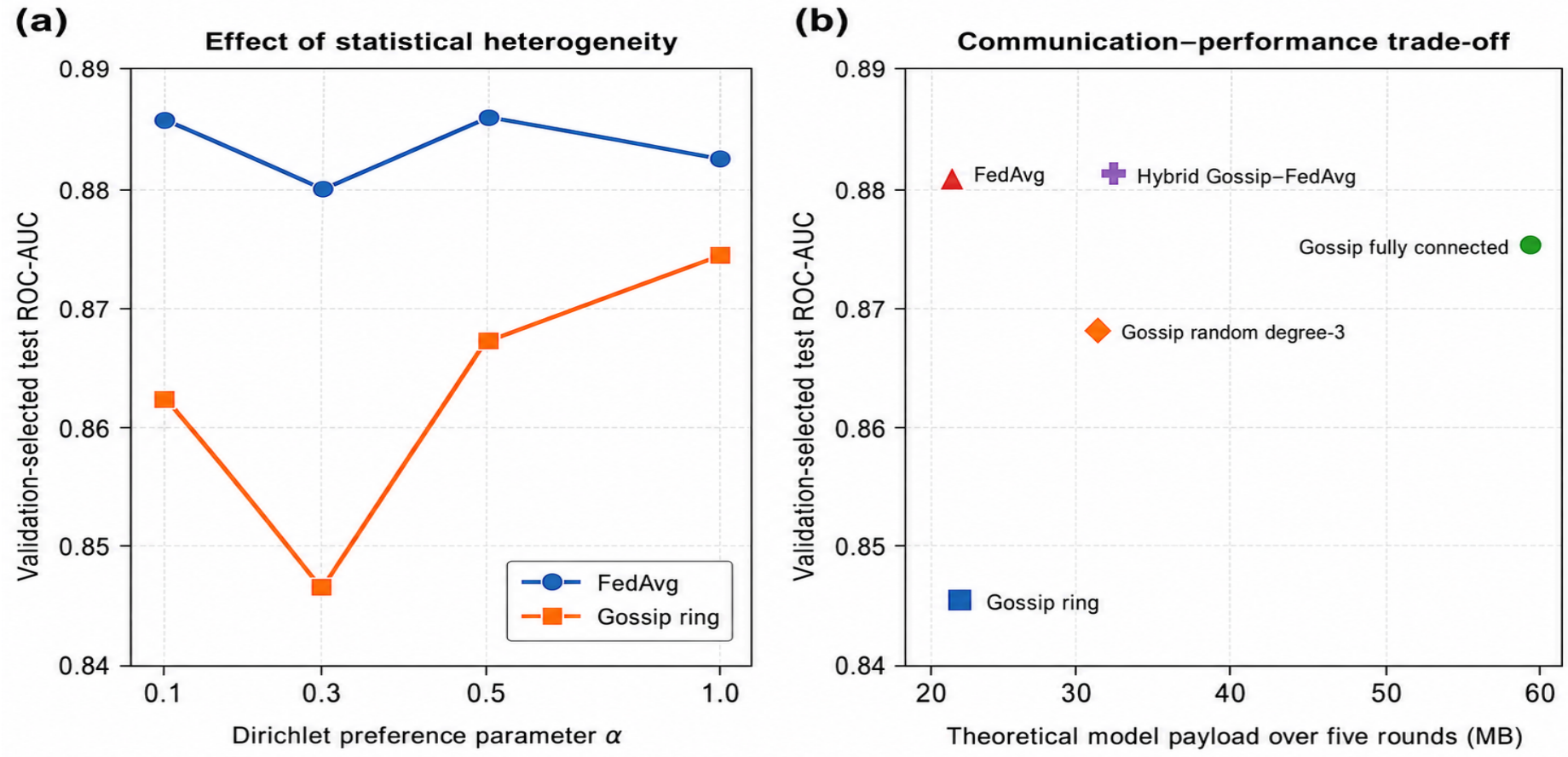

**Fig. 5** Effects of statistical heterogeneity and communication structure: (a) validation-selected test ROC-AUC of FedAvg and ring gossip across $\alpha \in \{0.1, 0.3, 0.5, 1.0\}$, and (b) communication–performance trade-off under the $\alpha = 0.3$ allocation

Ring-gossip stability was additionally examined through mixing-coefficient and learning-rate sensitivity analyses. With the baseline learning rate of $1 \times 10^{-3}$, test ROC-AUC values were 0.8012, 0.8065, 0.8455, and 0.8425 for $\beta = 0.25$, 0.50, 0.6667, and 1.00, respectively. The strongest result was therefore obtained at $\beta = 0.6667$, corresponding to equal weighting of the locally updated model and the two neighboring models in the ring. Learning rate also had a marked effect: reducing the rate from $1 \times 10^{-3}$to $5 \times 10^{-4}$ increased ring-gossip ROC-AUC from 0.8455 to 0.8770, while a further reduction to $1 \times 10^{-4}$ produced ROC-AUC of 0.8666. These experiments were treated as protocol-sensitivity analyses and were not used to retrospectively replace the baseline settings in the primary cross-protocol comparison.

### 4.4. Independent repeated evaluation and reliability analysis

To examine robustness to patient splitting, client allocation, and model initialization, ring-topology gossip learning, FedAvg, and Hybrid Gossip–FedAvg were evaluated in three independent repetitions using seeds 42, 123, and 2026. The patient-level training, validation, and test partitions and the workload-balanced, Dirichlet-guided client allocation were regenerated for each repetition, while all three protocols within a repetition used identical partitions and node assignments. Checkpoints were selected using validation ROC-AUC, and clinically oriented decision thresholds were determined exclusively on the corresponding validation partition before being applied unchanged to the held-out test partition. Table 3 reports the mean and standard deviation across the three repetitions. The primary single-partition experiments and the independent repeated analysis were generated from different patient splits and initialization settings. Their absolute performance levels are therefore not directly compared; instead, the repeated analysis is used to assess robustness and reliability across regenerated partitions.

FedAvg and Hybrid Gossip–FedAvg obtained the same mean ROC-AUC when rounded to four decimal places, while Hybrid achieved the numerically highest mean PR-AUC. Their differences were small relative to the observed variation across repetitions and therefore do not support a claim of statistical superiority. Ring gossip produced a slightly lower mean ROC-AUC but a comparable PR-AUC, indicating that its main limitation was in overall ranking discrimination rather than positive-class precision–recall behavior. FedAvg achieved the lowest mean Brier score, whereas Hybrid obtained the lowest mean ECE. Because the ECE standard deviations were comparatively large, particularly for Hybrid and ring gossip, the calibration results should be interpreted descriptively rather than as definitive evidence of protocol ranking.

**Table 3** Independent three-repetition discrimination, calibration, clinical operating-point, and patient-level burden results under the $\alpha = 0.3$ workload-balanced, Dirichlet-guided allocation. Values are reported as mean $\pm$ standard deviation. Clinical thresholds were selected on the validation partition and applied unchanged to the held-out test partition.

| ***Learning protocol*** | ***ROC-AUC*** | ***PR-AUC*** | ***Brier score ↓*** | ***ECE ↓*** | ***Spec. at 95% sen.*** | ***Sen. at 95% spec.*** | ***Burden MAE ↓*** | ***Burden Pearson r ↑*** |
|---|---|---|---|---|---|---|---|---|
| *Gossip learning, ring* | 0.9018 ± 0.0080 | 0.8203 ± 0.0152 | 0.1516 ± 0.0255 | 0.1539 ± 0.0926 | 0.561 ± 0.086 | 0.555 ± 0.042 | 0.175 ± 0.034 | 0.640 ± 0.094 |
| *Federated Averaging* | 0.9082 ± 0.0037 | 0.8219 ± 0.0211 | 0.1335 ± 0.0163 | 0.1219 ± 0.0512 | 0.627 ± 0.074 | 0.574 ± 0.030 | 0.148 ± 0.025 | 0.690 ± 0.089 |
| *Hybrid Gossip–FedAvg* | 0.9082 ± 0.0043 | 0.8240 ± 0.0211 | 0.1342 ± 0.0251 | 0.1126 ± 0.0831 | 0.614 ± 0.080 | 0.569 ± 0.036 | 0.149 ± 0.039 | 0.681 ± 0.098 |

At the operating point targeting 95% sensitivity, FedAvg achieved the highest mean specificity of 0.627, followed by Hybrid Gossip–FedAvg at 0.614 and ring gossip at 0.561. At the threshold targeting 95% specificity, the corresponding mean sensitivities were 0.574, 0.569, and 0.555. Thus, FedAvg and Hybrid showed closely comparable clinically oriented behavior, while ring gossip was somewhat less favorable at both operating points. Importantly, these values were obtained without any test-set threshold optimization.

Patient-level IDC burden estimation showed a similar pattern. FedAvg achieved the lowest burden MAE ($0.148 \pm 0.025$) and the highest Pearson correlation ($0.690 \pm 0.089$), followed closely by Hybrid Gossip–FedAvg with MAE of $0.149 \pm 0.039$ and Pearson correlation of $0.681 \pm 0.098$. Ring gossip produced a higher burden MAE ($0.175 \pm 0.034$) and a lower Pearson correlation ($0.640 \pm 0.094$). Complementary burden metrics were consistent with this pattern. Additionally, RMSE values were $0.219 \pm 0.038$, $0.186 \pm 0.027$, and $0.188 \pm 0.042$ for ring gossip, FedAvg, and Hybrid Gossip–FedAvg, respectively, while the corresponding Spearman correlations were $0.628 \pm 0.069$, $0.667 \pm 0.061$, and $0.662 \pm 0.066$. These findings indicate that periodic or round-wise global averaging was associated with more stable patient-level burden estimation than sparse peer-to-peer diffusion alone, although Hybrid remained closely aligned with FedAvg.

### 4.5. Backbone robustness and extended-budget FedAvg analysis

To examine whether the ring-gossip findings were specific to the lightweight CNN architecture, a limited backbone robustness experiment was first conducted using ResNet-18 under the principal $\alpha = 0.3$ patient partition and ring-topology communication setting. The lightweight CNN achieved a validation-selected test ROC-AUC of 0.8455 and PR-AUC of 0.5820 at round 4, whereas ResNet-18 achieved a ROC-AUC of 0.8606 and PR-AUC of 0.6204 at round 5. Thus, the ResNet-18 backbone increased ROC-AUC by 0.0151 and PR-AUC by 0.0384 in this matched ring-gossip comparison, indicating that the observed decentralized-learning behaviour was not confined to the lightweight CNN.

A second auxiliary experiment evaluated FedAvg with a ResNet-18 backbone trained from scratch under an extended maximum budget of 20 communication rounds. Each round comprised one complete local epoch at each of the six clients, corresponding to approximately 238–253 mini-batches per client and 1,500 local mini-batch updates across all clients per round. Validation-based early stopping terminated training after 16 completed rounds and selected the checkpoint from round 13, which achieved a validation ROC-AUC of 0.9292. Evaluation of the held-out test partition at this checkpoint yielded a ROC-AUC of 0.9118, PR-AUC of 0.8065, accuracy of 0.8091, sensitivity of 0.8901, specificity of 0.7758, and Brier score of 0.1354. The test partition was evaluated only once after restoration of the validation-selected checkpoint.

The extended ResNet-18 experiment demonstrates that a higher-capacity backbone combined with a longer validation-controlled aggregation schedule can substantially improve absolute FedAvg discrimination. Relative to the five-round lightweight FedAvg benchmark, ROC-AUC increased from 0.8801 to 0.9118 and PR-AUC from 0.6725 to 0.8065. This improvement was accompanied by a substantial increase in communication requirements. The ResNet-18 model state contained approximately 11.19 million float32 values, corresponding to 44.75 MB per model transmission and 536.96 MB of theoretical FedAvg payload per communication round. The 16 completed rounds therefore represented approximately 8.59 GB of theoretical model traffic, compared with 22.63 MB for the five-round lightweight FedAvg experiment. These results show that stronger backbones and longer aggregation schedules can improve predictive performance, but the gain must be considered together with the associated increase in model-transmission cost.

## 5. Discussion

This study compared server-based FedAvg, fully decentralized gossip learning, and Hybrid Gossip–FedAvg for IDC patch classification under patient-disjoint, workload-balanced, Dirichlet-guided data allocation. The principal single-partition experiment showed closely matched discrimination for Hybrid Gossip–FedAvg and FedAvg, with ROC-AUC values of 0.8811 and 0.8801, respectively. The independent three-repetition analysis produced the same mean ROC-AUC of 0.9082 for both protocols when rounded to four decimal places, while their PR-AUC, calibration, clinical operating-point, and patient-level burden results also remained close. Accordingly, the patient-level burden analysis should be interpreted as a clinically motivated aggregation study rather than as a full slide-level or patient-level diagnostic system. These findings do not establish a statistically superior protocol; rather, they indicate that periodic global aggregation can preserve most of the stability of FedAvg while allowing peer-to-peer model diffusion between synchronization rounds. The gossip-only results demonstrated that decentralized performance depended strongly on communication topology and local optimization settings. Under the principal $\alpha = 0.3$ allocation, ROC-AUC increased from 0.8455 with ring gossip to 0.8678 with random degree-3 gossip and 0.8751 with fully connected gossip, but the corresponding theoretical model payload also increased from 22.63 to 56.56 MB. This trend is consistent with the expectation that denser graphs accelerate information propagation and improve agreement, while increasing communication requirements [23,37]. However, connectivity alone did not determine performance. Ring-gossip ROC-AUC increased from 0.8455 to 0.8770 when the learning rate was reduced from $1 \times 10^{-3}$ to $5 \times 10^{-4}$, and the mixing-coefficient experiment showed that weak neighbor exchange produced substantially poorer results. Gossip learning should therefore be interpreted as a coupled optimization–communication process in which graph structure, learning rate, local data heterogeneity, and mixing strength jointly influence convergence.

The auxiliary ResNet-18 analyses further demonstrated that backbone capacity and communication-round budget materially influenced the achievable performance. Under the matched five-round ring-gossip setting, ResNet-18 increased ROC-AUC from 0.8455 to 0.8606 and PR-AUC from 0.5820 to 0.6204 relative to the lightweight CNN. In the extended FedAvg experiment, the validation-selected round-13 ResNet-18 checkpoint achieved a test ROC-AUC of 0.9118 and PR-AUC of 0.8065, substantially exceeding the corresponding five-round lightweight FedAvg values of 0.8801 and 0.6725. These gains were accompanied by approximately 8.59 GB of theoretical model traffic over the 16 completed rounds. The results therefore show that stronger backbones and longer aggregation schedules can produce marked improvements in discrimination, while also introducing substantially greater model-transmission requirements.

The round-wise consistency analysis further clarified this behavior. Relative parameter drift, node-level probability dispersion, and non-unanimous prediction rates generally decreased over successive ring-gossip rounds, indicating increasing agreement among the node models. Nevertheless, validation ROC-AUC did not improve monotonically and declined slightly after its selected checkpoint. Thus, increased consensus does not necessarily imply improved generalization: node models may become more similar while simultaneously converging toward a less favorable region of the parameter space. Validation-based checkpoint selection is therefore important for decentralized learning, particularly when only a small number of communication rounds are available and local datasets are non-IID. Calibration and clinically oriented operating points revealed additional differences that were not apparent from ROC-AUC alone. In the independent repetitions, FedAvg obtained the lowest mean Brier score, whereas Hybrid Gossip–FedAvg produced the lowest mean ECE. Both protocols also achieved higher specificity at 95% sensitivity and higher sensitivity at 95% specificity than ring gossip. The margins between FedAvg and Hybrid were small, while calibration variability across repetitions was comparatively large. These results emphasize that discrimination and calibration should be evaluated separately and that operating thresholds must be selected on validation data rather than optimized on the held-out test set. Patient-level IDC burden analysis produced a similar pattern: FedAvg and Hybrid showed nearly identical MAE, RMSE, and correlation values, whereas ring gossip was less accurate. Periodic or round-wise global aggregation therefore appeared beneficial for stabilizing probability estimates that were subsequently averaged across patches.

From a system perspective, each protocol provides a different compromise. FedAvg achieved strong discrimination, calibration, and burden estimation with the same theoretical model payload as ring gossip, but it required a central server to be available during every communication round. Fully connected gossip removed this server dependency but required substantially more peer-to-peer traffic. Random degree-3 gossip provided an intermediate decentralized option, while Hybrid Gossip–FedAvg reduced the frequency of server aggregation and achieved performance close to FedAvg with an intermediate communication payload. These observations should not be interpreted as evidence of fault tolerance or privacy preservation, because node failures, dropped communication links, asynchronous participation, adversarial updates, and privacy attacks were not evaluated. Maintaining raw images locally reduces direct data transfer but does not provide a formal privacy guarantee for exchanged model parameters [26,30,31].

Several limitations remain. The experiments used a single publicly available patch dataset, one principal lightweight CNN, only five scheduled communication rounds, and six simulated clients rather than data collected prospectively from distinct institutions. Although the independent analysis included three patient-level repetitions, this number is insufficient for strong statistical inference, and the broader topology, heterogeneity, learning-rate, and mixing-coefficient experiments were conducted on a fixed principal partition. The auxiliary ResNet-18 experiments were protocol-specific and used different optimization and communication-round budgets. They nevertheless strengthen the evidence that the reported findings are not confined to the lightweight CNN and provide an additional analysis of the performance–communication trade-off associated with scaling model capacity, although they do not constitute a fully matched backbone comparison across all three protocols. The $50 \times 50$ patches do not preserve whole-slide spatial context, stain normalization and domain-shift correction were not investigated, and the patient identifiers provided in the released dataset were used as grouping units without access to additional clinical metadata. Communication values represent theoretical float32 model payloads and exclude protocol overhead, serialization, retransmission, and congestion effects.

Future work should evaluate the protocols on external multi-institutional datasets, larger pathology encoders, stain-normalized and multiscale inputs, slide-level aggregation, and longer communication schedules. Additional studies should examine asynchronous gossip, client dropout, topology adaptation, compressed or quantized model exchange, secure aggregation, differential privacy, and alternative Hybrid synchronization intervals. Within the scope of the present experiments, FedAvg provided the most consistently reliable server-based baseline, topology-aware gossip offered a viable fully decentralized alternative when communication structure and optimization were carefully controlled, and Hybrid Gossip–FedAvg provided a practical middle ground between continuous global coordination and purely peer-to-peer training.

## 6. Conclusion

This study presented a controlled comparison of server-based FedAvg, fully decentralized gossip learning, and Hybrid Gossip–FedAvg for IDC histopathology patch classification under patient-disjoint, workload-balanced, Dirichlet-guided data allocation. The evaluation jointly considered predictive discrimination, calibration, clinically motivated operating points, patient-level IDC burden, node-level consistency, communication topology, and theoretical model payload. Ring, random degree-3, and fully connected gossip graphs were examined together with sensitivity analyses for the heterogeneity-control parameter, gossip mixing coefficient, and learning rate. In the principal $\alpha = 0.3$ experiment, Hybrid Gossip–FedAvg and FedAvg achieved closely matched test ROC-AUC values of 0.8811 and 0.8801, respectively, while fully connected gossip reached 0.8751. The three-repetition robustness analysis produced the same mean ROC-AUC of 0.9082 for FedAvg and Hybrid Gossip–FedAvg when rounded to four decimal places. FedAvg achieved the lowest mean Brier score and slightly stronger clinical operating-point and burden results, whereas Hybrid achieved the lowest mean ECE. These small differences do not support a claim of statistical superiority but indicate that periodic global aggregation can retain much of the stability of FedAvg while reducing the frequency of server coordination. The auxiliary ResNet-18 analyses further showed that stronger backbones and longer aggregation schedules can markedly improve absolute discrimination, although these gains were accompanied by substantially higher model-transmission requirements.

The gossip-only results revealed a clear communication–performance trade-off. Denser communication graphs improved ROC-AUC but increased transmitted model payload, while ring gossip remained sensitive to the learning rate and mixing coefficient. Within the scope of the present experiments, FedAvg represents the most consistently reliable option when continuous server availability is acceptable, topology-aware gossip provides a fully decentralized alternative when communication structure and optimization are carefully controlled, and Hybrid Gossip–FedAvg offers an intermediate design combining peer-to-peer diffusion with periodic global synchronization. These findings establish a reproducible benchmark for future studies of decentralized and hybrid learning in computational pathology.